\documentclass[runningheads]{llncs}

\usepackage[T1]{fontenc}
\def\doi#1{\href{https://doi.org/\detokenize{#1}}{\url{https://doi.org/\detokenize{#1}}}}
\usepackage{graphicx}
\usepackage{listings}
\usepackage{multirow}

\begin{document}
\title{Automatic Quality Assessment of First Trimester Crown-Rump-Length Ultrasound Images}

\author{Sevim Cengiz\inst{1}\orcidID{0000-0002-1332-3007} 
Ibraheem Hamdi \inst{1}\orcidID{0000-0002-5045-2238}
Mohammad Yaqub\inst{1}\orcidID{0000-0001-6896-1105}}
%

%
\institute{%
  Computer Vision, Mohamed bin Zayed University of Artificial Intelligence,\\
  Abu Dhabi, United Arab Emirates\\
  \email{sevim.cengiz@mbzuai.ac.ae, ibraheem.hamdi@mbzuai.ac.ae, mohammad.yaqub@mbzuai.ac.ae}\\
  \url{https://mbzuai.ac.ae/biomedia}%
}

\maketitle              


\begin{abstract}
Fetal gestational age (GA) is vital clinical information that is estimated during pregnancy in order to assess fetal growth. This is usually performed by measuring the crown-rump-length (CRL) on an ultrasound image in the Dating scan which is then correlated with fetal age and growth trajectory. A major issue when performing the CRL measurement is ensuring that the image is acquired at the correct view, otherwise it could be misleading. Although clinical guidelines specify the criteria for the correct CRL view, sonographers may not regularly adhere to such rules. In this paper, we propose a new deep learning-based solution that is able to verify the adherence of a CRL image to clinical guidelines in order to assess image quality and facilitate accurate estimation of GA. We first segment out important fetal structures then use the localized structures to perform a clinically-guided mapping that verifies the adherence of criteria.  The segmentation method combines the benefits of Convolutional Neural Network (CNN) and the Vision Transformer (ViT) to segment fetal structures in ultrasound images and localize important fetal landmarks. For segmentation purposes, we compare our proposed work with UNet and show that our CNN/ViT-based method outperforms an optimized version of UNet. Furthermore, we compare the output of the mapping with classification CNNs when assessing the clinical criteria and the overall acceptability of CRL images. We show that the proposed mapping is not only explainable but also more accurate than the best performing classification CNNs.

\end{abstract}
\section{Introduction}

Fetal ultrasound is important to help sonographers assess fetal growth and check for abnormalities. A Dating scan is typically performed between 11 - 13 weeks of gestation. During this scan, a crown-rump-length (CRL) view image must be acquired to measure the fetal length, which is taken from the top of the head to the bottom of the rump. CRL is a critical biometric measurement that can be used to calculate GA. Precise CRL measurement is key for detecting potential fetal anomalies such as small for gestational age (SGA) and large for gestational age (LGA) where small or large indicate deviations from normal fetal size with respect to the number of weeks of the pregnancy. To perform the correct measurement, the CRL view must satisfy the clinical guidelines listed and described in Table 1 \cite{salomon,NHS1,NHS2}. However, acquiring the correct fetal ultrasound view is a challenging task due to different factors such as sonographer experience, fetal position, and maternal characteristics \cite{Yaqub2021}. In the CRL view, the fetus might be at a hyperextended which represents the extensive gap between the chin and chest, or hypoflexed position which represents the narrow gap between the chin and chest, leading to an incorrect CRL measurement and consequently an incorrect GA estimation. In addition, since quality of the acquired ultrasound is highly dependent on sonographer ability, there is no guarantee that the acquired images adhere to the clinical guidelines. For instance, authors in \cite{Yaqub2018} suggested that a large number of fetal ultrasound anomaly scans do not adhere to clinical guidelines when a large set of scans in a well-established maternity unit were manually assessed. Manual auditing to check the adherence of acquired CRL images to clinical guidelines could be performed to alleviate this issue but it is not real-time and lacks instant feedback to the sonographer. Since manual auditing is expensive, time-consuming, and typically performed on a small percentage of scans \cite{expensive}, it is necessary to find an efficient solution for such a challenging problem.

Fetal ultrasound image analysis is gaining a lot of interest. Most of the published work for Dating scan image analysis focused on the placenta. For instance, Yang et al. \cite{yang2019} conducted a study about a fully automatic framework to segment various anatomical structures such as fetus, gestational sac, and placenta. Similarly, Zimmer et al. \cite{zimmer2019} used CNN to extract and segment the entire placenta and estimate its volume during the last trimester of pregnancy using ultrasound images. An improved version of this solution which incorporates the placental position to improve segmentation accuracy was proposed \cite{zimmer2020}. Looney et al. \cite{looney2018} introduced OxNNet model to segment the placenta automatically. Recently, OxNNet was improved to segment amniotic fluid and fetus as well using a multi class CNN \cite{looney2021}. 

Little work has been published to-date on fetal CRL segmentation and assessment. Ryou et al. \cite{ryou2019} has developed a 3D CNN which is able to segment fetal structures at CRL view and perform the CRL measurement. Although the authors analyzed 3D ultrasound, the generated CRL view was assumed to adhere to the clinical guidelines. Similarly, Cengiz et al. \cite{cengiz2021} proposed a CNN-based method to segment the body and head of the fetus from ultrasound images in order to measure the CRL and subsequently estimate the gestational age (GA). These authors have also assumed that the images were acquired at the correct CRL view, which is not necessarily correct unless verified. 

Transformer is a unique architecture that has inspired a lot of interest in the computer vision (CV) community despite being initially built for sequence-to-sequence modeling in natural language processing (NLP). Vision Transformer (ViT) \cite{vit} is proposed as the first image recognition model that is solely based on Transformer by achieving equivalent performance to other state-of-the-art (SOTA) convolution-based approaches. The self-attention mechanism of Transformer is powerful at modeling global contextual information \cite{transunet}. On the other hand, CNNs typically concentrate on a relatively small neighborhood in the entire image, losing global context and having weak performance when forming global dependenciess\cite{transunet}. Therefore, there is a need for a method that can acquire in-depth localization information at all network stages and transfer the context over a long distance within the network. A unique architecture called TransUNet combines Transformer and UNet for this purpose. Authors \cite{transunet} claimed that their model surpassed other methods when tested on cardiac and multi-organ segmentation.

\begin{figure}[t]
\centering
\includegraphics[width=5.0in]{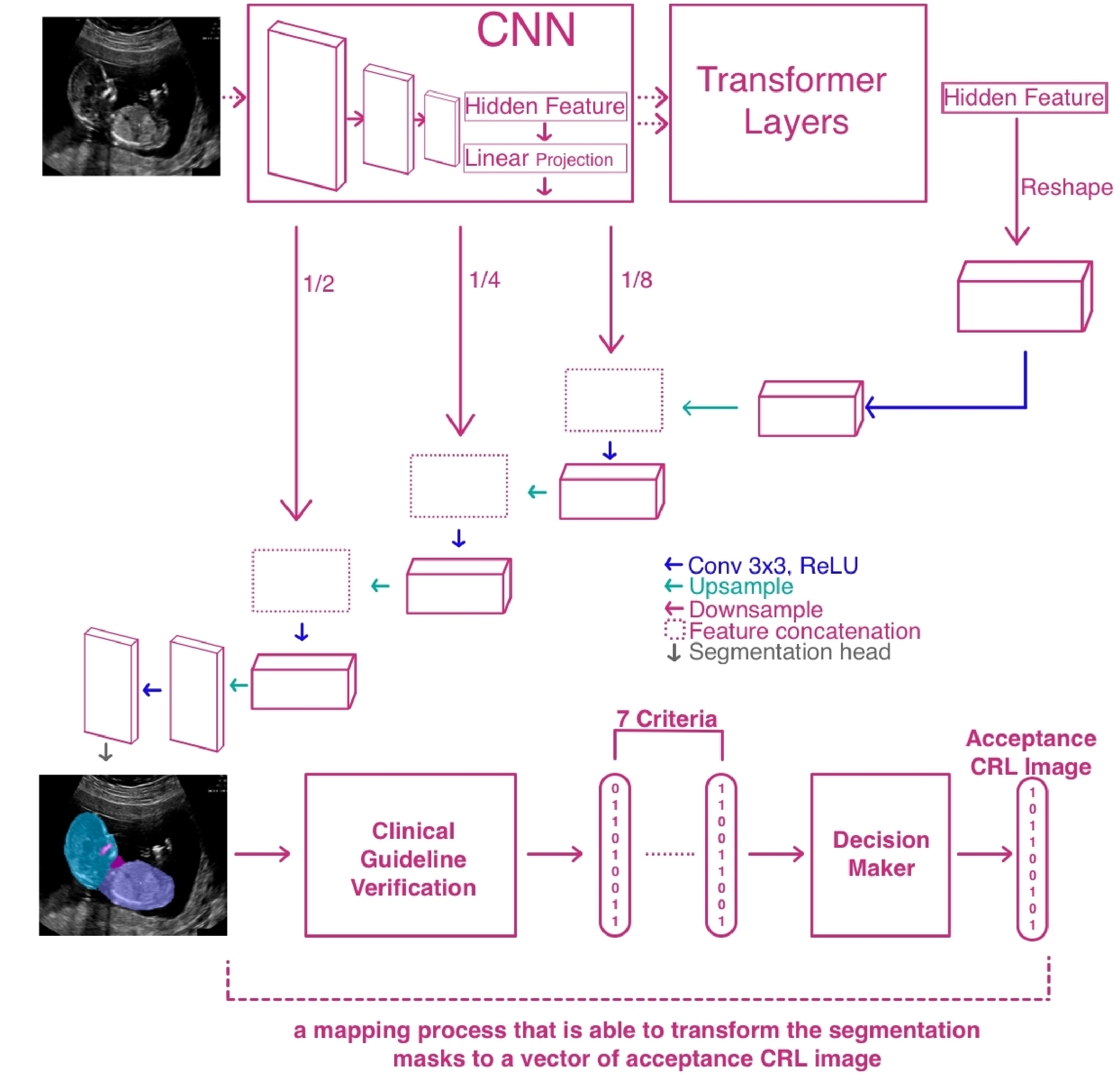}
\caption{Flowchart of the proposed method with Fetal TransUNet to verify the adherence to clinical guidelines of the fetal ultrasound images.}
\label{fig:prospoedsolution}
\end{figure}

In this study, a segmentation-guided heuristic model is developed by proposing 1) a ViT+UNet model to segment the fetal ultrasound structures and 2) a mapping process from an ultrasound image and a segmentation mask to a quality assessment that checks for image adherence to clinical guidelines. Figure \ref{fig:prospoedsolution} shows the flowchart for the proposed solution. This paper does not aim to propose a new algorithm nor expand on an existing one but rather introduce a complete solution for this challenging and unexplored problem. The contributions of this work are three folds:
\begin{enumerate}
\item To our knowledge, this is first attempt for automatic fetal CRL quality assessment.
\item The method utilizes the benefit of a ViT and UNet to ensure accurate fetal segmentation results.
\item The mapping from fetal segmentation to clinical criteria not only outperforms CNN-based approaches but also provides explainable results which make the solution more adaptable for clinical use.
\end{enumerate}

\section{Materials and Methods} 
\subsection{Dataset} 

696 fetal CRL ultrasound images from the first trimester were extracted out of the hospital archive, excluding twin pregnancies. An expert segmented 4 different structures using an annotation tool (ITK-Snap \cite{itksnap}). These structures are the head, body, fetal palate, and the gap between the chin and chest. After manual segmentation, another experienced researcher examined all the segmentation masks and made corrections if needed. 

Some fetuses were not scanned in a neutral position, meaning they were not at the mid-sagittal plane, resulting in loss of view of the fetal palate. In these cases, the image does not have a segmented region representing the fetal palate. Similarly, if the fetus was at a hyperflexed (over-stretched) or hypoflexed (under-stretched) position, there is either a large or little to no gap present between the chin and chest, respectively. For each CRL image, a segmentation mask and ground truth for the 7 criteria (a vector of 7 binary values) have been generated. In clinical practice \cite{wanyonyi2014}, if the sum of the 7 criteria is greater than 3, then the image is considered clinically acceptable. Therefore, we used this condition to decide whether an image is acceptable for CRL measurement or not. 

\subsection{Localization of Fetal Structures}

Although UNet \cite{ronneberger} has shown great success in medical image segmentation, Vision Transformer \cite{vistrans} has shown successful results in some natural image segmentation problems \cite{segmenter}. CNNs are typically good at capturing features in local neighborhood, while ViT models are able to capture a global representation of images\cite{transunet}. Recently, researchers have investigated the aggregation of both CNNs and ViT for different tasks. For instance, Transformer and UNet (TransUNet) \cite{transunet} have been combined to perform multi-organ segmentation in CT scans. Results show that TransUNet has higher metric scores and better performance when compared to UNet or ViT separately \cite{transunet}. 

In this study, we adopted TransUNet (Fetal TransUNet) and UNet for the problem of fetal structure segmentation. UNet is a multi class segmentation model that has a fully convolutional encoder-decoder architecture with 4 blocks, each of them consisting of convolutional and ReLu layers with max pooling. Fetal TransUNet consists of a CNN that learns a feature representation for an input image. This helps the CNN learn discriminative feature vector. The feature vector generated from the CNN is fed to a set of transformer layers which allows for a global encoding of the feature vector. To decode the hidden feature and produce the final segmentation mask, a cascaded upsampler comprised of a number of upsampling stages was used. Cascaded upsampler and CNN-Transformer hybrid encoder form a u-shaped architecture that allows for feature aggregation at multiple resolution levels via skip-connections. 

Due to the small amount of available data in this work, 3-fold cross validation was applied. We have experimented with multiple hyper-parameters for UNet and Fetal TransUNet. With 100 epochs, we have experimentally chosen learning rate, batch size, and image size as 10e(-5), 24, and 224, respectively. In both implementations, random rotation (degrees [-10, 10]) and horizontal flipping were applied as data augmentations.

\subsection{Criteria Assessment for Optimal CRL Plane}

The Fetal Anomaly Screening Programme (FASP) guideline \cite{NHS1,NHS2} recommends a set of criteria to be checked before CRL measurement is taken to ensure that the image is acquired at a correct view (see Table \ref{table-setcriteria}). In this AI-based heuristic method, TransUNet and UNet are implemented to produce segmented fetal body structures from ultrasound images, and then a mapping is developed to evaluate the adherence to clinical criteria by providing explainable results on those masks. Neutral position, horizontal orientation, fetal palate, magnification, fetal face direction, left and right caliper definitions are the criteria investigated by considering the geometrical and analytical aspects while creating a spreadsheet to represent the criteria acceptance with 1s and 0s (step by step evaluation details for each criterion are given in Table \ref{table-setcriteria}). More details are given in the supplementary video. The overall image plane acceptance for the CRL measurement was then determined by checking whether the total score is more than 3 out of 7. \cite{wanyonyi2014}. The spreadsheets were then used to calculate accuracy, recall, precision, and F1 scores for both methods.GitHub repository of the proposed method will be public after the acceptance of the paper. 

\begin{table}[t]
\caption{Image assessment criteria for the CRL measurement}
\centering
\resizebox{\columnwidth}{!}{
\begin{tabular}{|c|l|}
\hline
\textbf{Image Guidance Criteria} & \multicolumn{1}{c|}{\textbf{Description and Assessment in the Proposed Model}}                                                                                                                                                                                                                             \\ \hline
1-Neutral Position               & \begin{tabular}[c]{@{}l@{}}Body in a neutral position without hyperflexion or extension. The gap region \\ between the chin and chest which represents hyperflexion or extension was analyzed\\ automatically by using the segmentation results of UNet or TransUNet to understand \\how the region is small or large. 
.\end{tabular}                                                                                                      \\ \hline
2-Horizontal Orientation         & \begin{tabular}[c]{@{}l@{}}The angle of the CRL axis should be between ±15 degrees to the horizontal line. \\ CRL line was measured from the segmented images based on the method in \\ Cengiz et al. {[}14{]}. The slope between the horizontal axis line and the CRL line\\ was calculated automatically.\end{tabular} \\ \hline
3-Fetal Palate                   & \begin{tabular}[c]{@{}l@{}}Head should be in mid-sagittal view with a clear fetal palate. Presence of palate \\ was used to confirm that fetus is at mid-sagittal.\end{tabular}                                                                                                                            \\ \hline
4-Magnification                  & \begin{tabular}[c]{@{}l@{}}The horizontal projection of the CRL line was checked automatically to make sure \\ that the whole CRL should take up more than 60\% of the image. \end{tabular}                                                                                                                                                                                                                                                   \\ \hline
5-Left definition for caliper    & \begin{tabular}[c]{@{}l@{}}Clearly defined crown and rump for correct caliper placement. Pixel intensities \\ in 20x20 size square areas from the first and last points of CRL line were evaluated\end{tabular}                                                                                             \\ \cline{1-1}
6-Right definition for caliper   & to ensure whether the head and bottom points of the body are clearly visible.                                                                                                                                                                                                                              \\ \hline
7-Direction of Fetal Face        & \begin{tabular}[c]{@{}l@{}}Fetal face should be looking up. Automatically identified whether the fetus’ face was looking \\ up or down by the place of a perpendicular line from the intersection point \\ between the head and body to the CRL line.\end{tabular}                                                       \\ \hline
\end{tabular}
}
\label{table-setcriteria}
\end{table}

\subsection{Experiments}

To compare our solution with CNN-based networks, DenseNet-121 and ResNet-34 were adopted to classify each criterion directly from the fetal ultrasound images as well as the clinical acceptance of the CRL measurement. The spreadsheet with the predicted labels for each criterion was compared to the results of the proposed heuristic model and the accuracy, recall, precision, and F1 scores were calculated. Both models used in our experiments were pretrained on ImageNet to speed up convergence. In addition, weighted cross-entropy loss and Adam optimizer as well as and 3-fold cross validation were used to optimize results. After experimenting with different combinations of hyperparameters, a learning rate of 0.003 and 0.01, batch size of 32, and 24, and image size of 256 were chosen for UNet and TransUNet, respectively. The models were trained for 100 epochs.  

\section{Results}

Figure \ref{fig2} shows the examples of clinically accepted (the first row), and rejected (the second row) CRL images for each criterion. The third row shows how deep learning and mapping-based image assessment for CRL measurement are done bu using the segmentation masks at each criterion.

\begin{table}[t]
\caption{Mean and standard deviation of Dice, Jaccard, precision, recall scores of UNet and Fetal TransUNet segmentations}
\centering
\resizebox{0.8\columnwidth}{!}{%
\begin{tabular}{|c|c|l|l|l|l|}
\hline
\multicolumn{1}{|l|}{} & \multicolumn{1}{l|}{} & \textbf{Dice} & \textbf{Jaccard} & \textbf{Precision} & \textbf{Recall} \\ \hline
\multirow{2}{*}{\textbf{Head}} & UNet & 0.913±0.11 & 0.854±0.13 & 0.911±0.10 & 0.929±0.12 \\ \cline{2-6} 
                             & Fetal TransUNet & \textbf{0.927±0.06} & \textbf{0.869±0.08} & \textbf{0.932±0.06} & \textbf{0.929±0.07} \\ \hline
\multirow{2}{*}{\textbf{Body}} & UNet & 0.920±0.08 & 0.861±0.11 & 0.935±0.07 & 0.916±0.10 \\ \cline{2-6} 
                             & Fetal TransUNet & \textbf{0.927±0.06} & \textbf{0.868±0.07} & \textbf{0.940±0.05} & \textbf{0.918±0.07} \\ \hline
\multirow{2}{*}{\textbf{Fetal Palate}} & UNet & 0.382±0.35 & 0.299±0.29 & \textbf{0.349±0.42} & 0.464±0.34 \\ \cline{2-6} 
                                      & Fetal TransUNet & \textbf{0.398±0.36} & \textbf{0.315±0.29} & 0.340±0.40 & \textbf{0.573±0.34} \\ \hline
\multirow{2}{*}{\textbf{The gap}} & UNet & \textbf{0.546±0.30} & \textbf{0.432±0.26} & \textbf{0.571±0.35} & 0.618±0.29 \\ \cline{2-6} 
                                  & Fetal TransUNet & 0.536±0.29 & 0.418±0.25 & 0.538±0.33 & \textbf{0.656±0.28} \\ \hline
\end{tabular}%
}
\label{table-unet}
\end{table}

\begin{table}[]
\caption{Accuracy, recall, precision, and F1 score of DenseNet-121 (D.Net), ResNet-34 (R.Net), proposed model with UNet (PropU), and proposed model with Trans-UNet (PropTU).}
\resizebox{\columnwidth}{!}{
\begin{tabular}{|c|cccc|cccc|cccc|cccc|}
\hline
\multirow{2}{*}{} 
  & \multicolumn{4}{c|}{\textbf{Accuracy}} 
  & \multicolumn{4}{c|}{\textbf{Recall}} 
  & \multicolumn{4}{c|}{\textbf{Precision}} 
  & \multicolumn{4}{c|}{\textbf{F1}} \\ \cline{2-17} 
  & D.Net & R.Net & PropU & PropTU 
  & D.Net & R.Net & PropU & PropTU 
  & D.Net & R.Net & PropU & PropTU 
  & D.Net & R.Net & PropU & PropTU \\ \hline
Neutral position       & 52.1  & 53.0  & 50.1  & \textbf{53.5}  
                       & 58.1  & \textbf{69.2} & 63.6  & 64.3  
                       & \textbf{63.8}  & 61.8  & 51.8  & 61.9  
                       & 60.8  & \textbf{65.3} & 57.1  & 63.1 \\ \hline
Horizontal orientation & 80.7  & \textbf{94.0} & 91.5  & 92.0  
                       & 90.5  & 90.8  & \textbf{94.2} & \textbf{94.2}  
                       & 86.2  & 86.3  & 95.0  & \textbf{95.5}  
                       & 88.3  & 88.5  & 94.6  & \textbf{94.8} \\ \hline
Fetal palate           & 62.7  & 60.9  & 68.8  & \textbf{69.9}  
                       & \textbf{68.4} & 58.4  & 68.1  & 67.6  
                       & 50.6  & 62.8  & 77.5  & \textbf{82.8}  
                       & 58.2  & 60.5  & 72.5  & \textbf{74.4} \\ \hline
Magnification          & 74.2  & 64.7  & 92.8  & \textbf{94.8}  
                       & 77.2  & 58.8  & 92.3  & \textbf{96.4}  
                       & 87.3  & 90.9  & \textbf{98.5} & 96.4  
                       & 81.9  & 71.4  & 95.3  & \textbf{96.4} \\ \hline
Left caliper definition& 68.3  & \textbf{71.2} & 52.0  & 48.1  
                       & 83.4  & \textbf{88.5} & 77.6  & 76.0  
                       & \textbf{77.6} & 77.3  & 53.8  & 48.6  
                       & 80.4  & 82.5  & \textbf{63.5} & 59.3 \\ \hline
Right caliper definition& 63.6 & \textbf{66.3} & 65.0  & 62.2  
                       & 79.2  & \textbf{81.8} & 80.5  & 78.5  
                       & 74.4  & \textbf{75.5} & 71.3  & 69.4  
                       & 76.7  & \textbf{78.5} & 75.6  & 73.7 \\ \hline
Fetal face up/down     & 84.5  & 89.9  & 86.1  & \textbf{90.0}  
                       & 88.7  & \textbf{95.1} & 94.4  & 94.7  
                       & 94.3  & 94.2  & 90.5  & \textbf{94.7}  
                       & 91.4  & 94.6  & 92.4  & \textbf{94.7} \\ \hline
Acceptance for CRL     & \textbf{88.0} & 68.4  & 86.3  & 83.5  
                       & \textbf{94.7} & 71.0  & 93.7  & 94.2  
                       & \textbf{92.2} & 91.7  & 90.8  & 87.0  
                       & \textbf{93.4} & 80.0  & 92.2  & 90.5 \\ \hline
\end{tabular}
}
\label{table-cnn}
\end{table}

According to the spreadsheet manually prepared by the expert, the acceptance ratios of the neutral position, palate existence, magnification, the left definition of the caliper, the right definition of the caliper, fetal face direction, and horizontal angle were 0.64, 0.51, 0.75, 0.77, 0.76, 0.94, and 0.8,1 respectively. Table \ref{table-cnn} shows the accuracy, precision, recall, and F1 scores of DenseNet-121, ResNet-34, AI-based proposed method with UNet and TransUNet for the 7 clinical criteria as well as the acceptance of CRL measurement. Although both classifications and AI-based heuristic methods have their highest scores in horizontal orientation, magnification, fetal face direction, and acceptance of CRL measurement, neither managed to achieve more than 85\% for neutral position, fetal palate and left and right caliper definition. The accuracy of the DenseNet and ResNet based classification and the AI-heuristic model with UNet and TransUnet were 88.0\%, 68.4\%, 86.3\% and 83.5\%, respectively. Even if the scores in each criterion are close among the classification models, the proposed method based on the AI-heuristic approach with TransUnet has higher F1 scores for the criteria based evaluation.

\section{Discussion}
Results presented in Table \ref{table-unet} show that Fetal TransUNet had higher dice, jaccard, and recall scores for the fetal body structure segmentation including head, body, and fetal palate compared to UNet. In addition, the segmentation standard deviation of Fetal TransUNet is relatively smaller than UNet which suggests that our proposed method has a more consistent segmentation (i.e., fetal TranUNet is less likely to over- or under-segment).

No comparison was performed between the proposed Fetal TransUNet to a classical ViT since the original TransUNet paper \cite{transunet} has performed extensive evaluation against Vit, ViT with CNN models. Results suggest that TransUNet outperforms ViT and ViT with a CNN.
    
In the table \ref{table-cnn}, we showed that our proposed method with TransUNet has higher the F1 score compared to others, which is the most representative metric, among four out of seven criteria. However, our method with TransUnet has a slightly lower accuracy, recall, precision and F1 score than the DenseNet in terms of acceptance for CRL. On the other hand, the proposed model has great potential because it provides a reason for the output, which may assist in making a specific decision rather than the black-box approach used by CNNs. This heuristic method may contribute positively during the evaluation of an ultrasound image and result in a higher rate of acceptance for GA estimation. 

Horizontal orientation, magnification, fetal face direction, left and right definition of the caliper, as well as image acceptability for CRL measurement, were detected with high accuracy. However, the evaluation of the fetal palate and neutral positioning was more challenging considering the complexity of the images due to fetus movement, ambiguity of the region, and brightness and contrast of the ultrasound images.  In many scans, it was extremely difficult to visually determine neutral positioning even for a trained eye, hence it was not surprising that all methods produced their lowest accuracy in this criteria. Similarly, the presence of fetal palate in some scans was visually confusing since it blended with the chin or face, which resulted in lower evaluation scores. 

The novelty of this paper compared to the study conducted by \cite{cengiz2021} is that not only the network is composed of ViT+UNet but also creating from scratch a mapping process of the fetal segmentation images to a representative vector that shows the acceptability of the CRL image. In the mapping section, segmented masks were investigated taking into consideration the geometrical and analytical rules to verify and classify that it complies with a set of clinical criteria. As a result, instead of building traditional CNN classification models that would automatically record the predicted labels, the heuristic method combines the Fetal TransUNet with the mapping process that is the analytical assessment with geometrical aspects on output masks, showing an explainable and sensible method. Therefore, we believe this is the better solution for clinical use since the combined method allows us to investigate the sonographers assessment for the CRL measurement by making sure that the criteria guidelines are adhered to as well as the reason behind a specific decision rather than the black box approach of CNNs which is not capable of explaining that. This heuristic method may contribute positively during the assessment of an ultrasound image and result in a higher rate of acceptance for GA estimation. 

\section{Conclusion}
In this paper, we developed an AI-based heuristic method that segments the fetal ultrasound images and has a mapping process that is able to analyze the segmented images and its adherence to clinical criteria. We conclude that the use of a CNN to assess whether a CRL image adheres to a clinical criterion is not accurate nor explainable. On the other hand, our proposed method has a better accuracy and provides a reasonable explanation for each criterion when assessing the CRL image since it relies on localizing important structures (mimicking sonographer's assessment). 

The main limitation of this work is the small sized dataset. Therefore future work requires evaluation on a larger dataset. In addition, a major challenge we face in this work is label imbalance. More than 90\% of the images met 4 or more criteria and were tagged as acceptable for CRL measurement. Although it may seem that there is no novel technical contribution in the proposed method since we utilize the existing algorithm for ViT and UNet, to our knowledge this is the first publication aiming to address the challenging task of automatic quality assessment of the fetal CRL Dating images. For future work, to minimize human error in the ground truth labels, it would be best to rely on multiple sonographers to annotate and label the dataset.


\begin{thebibliography}{8}

\bibitem{salomon}
Salomon, L.J., et al.: ISUOG practice guidelines: performance of first-trimester fetal ultrasound scan. Ultrasound Obstet Gynecol. 2013 Jan;41(1):102-13. doi: 10.1002/uog.12342. Erratum in: Ultrasound Obstet Gynecol. 2013 Feb;41(2):240. PMID: 23280739.

\bibitem{NHS1}
Fetal anomaly screening programme: standards - GOV.UK., \url{https://www.gov.uk/government/publications/fetal-anomaly-screening-programme-standards}. Last accessed 26 Feb 2022.

\bibitem{NHS2}
Recommended criteria for measurement of fetal crown rump length (CRL) as part of combined screening for Trisomy 21 within the NHS in England, \url{http://www.pi.nhs.uk/ultrasound/CRL\_criteria\_for\_T21\_screening.pdf}. Last accessed 26 Feb 2022.

\bibitem{Yaqub2021}
Yaqub, M., Kelly, B., Noble, A., Papageorghiou, A.: The effect of maternal body mass index on fetal ultrasound image quality. American Journal of Obstretics \& Gynecology 225(2), 200-202 (2021). \doi10.1016/j.ajog.2021.04.248

\bibitem{Yaqub2018}
Yaqub, M., Kelly, B., Stobart, H., Napolitano, R., Noble, J.A., Papageorghiou, A.T.: Quality-improvement program for ultrasound-based fetal anatomy screening using large-scale clinical audit. Ultrasound Obstet Gynecol 2019;54:239–45.

\bibitem{yang2019}
Yang, X., et al.: Towards Automated Semantic Segmentation in Prenatal Volumetric Ultrasound. IEEE Trans. Med. Imaging 38(1), 180-193 (2019). \doi10.1109/TMI.2018.2858779. PMID: 30040635.

\bibitem{zimmer2019}
Zimmer, V.A., et al.: Towards Whole Placenta Segmentation at Late Gestation Using Multi-view Ultrasound Images. In: Shen D. et al. (eds.) MICCAI 2019. LNCS, vol 11768. Springer, Cham (2019). \doi10.1007/978-3-030-32254-0\_70

\bibitem{zimmer2020}
Zimmer, V.A., et al.: A Multi-task Approach Using Positional Information for Ultrasound Placenta Segmentation. In: Hu Y. et al. (eds.) ASMUS/PIPPI -2020. LNCS, vol 12437. Springer, Cham (2020). \doi10.1007/978-3-030-60334-2\_26

\bibitem{looney2018}
Looney, P., et al.: Fully automated, real-time 3D ultrasound segmentation to estimate first trimester placental volume using deep learning. JCI Insight. 3(11):e120178 (2018). \doi10.1172/jci.insight.120178. PMID: 29875312.

\bibitem{looney2021}
Looney, P., et al.: Fully Automated 3-D Ultrasound Segmentation of the Placenta, Amniotic Fluid, and Fetus for Early Pregnancy Assessment. IEEE Trans Ultrason Ferroelectr Freq Control. 68(6), 2038-2047 (2021). \doi10.1109/TUFFC.2021.3052143. PMID: 33460372.

\bibitem{schwartz2021}
Schwartz, N., et al.: Fully Automated Placental Volume Quantification From 3DUS for Prediction of Small-for-Gestational-Age Infants. J. Ultrasound Med. (2021). \doi10.1002/jum.15835. PMID: 34553780.

\bibitem{ryou2019}
Ryou H, Yaqub M, Cavallaro A, Papageorghiou AT, Alison Noble J. Automated 3D ultrasound image analysis for first trimester assessment of fetal health. Phys Med Biol. (2019) 64:185010. doi: 10.1088/1361-6560/ab3ad1

\bibitem{cengiz2021}
Cengiz, S., Yaqub, M.: Automatic Fetal Gestational Age Estimation from First Trimester Scans. In: Noble J.A., Aylward S., Grimwood A., Min Z., Lee SL., Hu Y. (eds.) Simplifying Medical Ultrasound. ASMUS -2021. LNCS, vol. 12967. Springer, Cham (2021). \doi10.1007/978-3-030-87583-1\_22

\bibitem{vit}
A. Dosovitskiy, L. Beyer, A. Kolesnikov, D. Weissenborn, X. Zhai, T. Unterthiner, M. Dehghani, M. Minderer, G. Heigold, S. Gelly et al., “An image is worth 16x16 words: Transformers for image recognition at scale,” arXiv preprint arXiv:2010.11929, 2020.

\bibitem{wanyonyi2014}
Wanyonyi, S.Z., Napolitano, R., Ohuma, E.O., Salomon, L.J. and Papageorghiou, A.T.: Image-scoring system for crown–rump length measurement. Ultrasound Obstet. Gynecol. 44(6), 649-654 (2014). \doi10.1002/uog.13376. PMID: 24677327.

\bibitem{ronneberger}
Ronneberger O., Fischer P., Brox T.: U-Net: Convolutional Networks for Biomedical Image Segmentation. In: Navab N., Hornegger J., Wells W., Frangi A. (eds.) MICCAI 2015. MICCAI 2015. LNCS, vol 9351, pp. 234-241. Springer, Cham (2015). \doi/10.1007/978-3-319-24574-4\_28

\bibitem{vistrans}
Dosovitskiy, A., et al.: An Image is Worth 16x16 Words: Transformers for Image Recognition at Scale (2020). arXiv preprint: arXiv:2010.11929. \doi10.48550/arXiv.2010.11929

\bibitem{segmenter}
Strudel, R., Garcia, R., Laptev, I., Schmid, C.: Segmenter: Transformer for Semantic Segmentation (2021). arXiv preprint: arXiv:2105.05633. \doi10.48550/arXiv.2105.05633

\bibitem{transunet}
Chen, J., et al.: TransUNet: Transformers Make Strong Encoders for Medical Image Segmentation (2021). arXiv preprint: arXiv:2102.04306

\bibitem{itksnap}
Yushkevich, P.A., et al.: User-guided 3D active contour segmentation of anatomical structures: Significantly improved efficiency and reliability. Neuroimage 2006 Jul 1;31(3):1116-28.

\bibitem{expensive}
Wang, S., Li, C., Wang, R. et al. Annotation-efficient deep learning for automatic medical image segmentation. Nat Commun 12, 5915 (2021). https://doi.org/10.1038/s41467-021-26216-9.

\end{thebibliography}
\end{document}